\documentclass[manuscript]{acmart}
\AtBeginDocument{%
  \providecommand\BibTeX{{%
    \normalfont B\kern-0.5em{\scshape i\kern-0.25em b}\kern-0.8em\TeX}}}

\setcopyright{acmcopyright}
\copyrightyear{2018}
\acmYear{2018}
\acmDOI{XXXXXXX.XXXXXXX}

\acmJournal{JACM}
\acmVolume{37}
\acmNumber{4}
\acmArticle{111}
\acmMonth{8}

\usepackage{multirow}

\usepackage{amsfonts}
\usepackage{threeparttable}
\usepackage{float}
\usepackage{enumitem}

\definecolor{color1}{RGB}{145,30,180}
\definecolor{color2}{RGB}{245,130,48}
\definecolor{color3}{RGB}{230,25,75}
\usepackage{subfigure}

\begin{document}

\title{Hyperbolic Hierarchical Knowledge Graph Embeddings for Link Prediction in Low Dimensions}

\author{Wenjie Zheng}
\email{wjzheng1996@gmail.com}
\affiliation{
  \institution{Nanjing University of Science and Technology}
  \country{China}
}

\authornote{Both authors contributed equally to this research.}
\author{Wenxue Wang}
\affiliation{
\institution{Key Laboratory of Intelligent Computing and Signal Processing, Ministry of Education, Anhui University}
\country{China}}
\authornotemark[1]

\author{Shu Zhao}
\affiliation{
\institution{Key Laboratory of Intelligent Computing and Signal Processing, Ministry of Education, Anhui University}
\country{China}}

\author{Fulan Qian}
\authornote{Corresponding Author.}
\authornotemark[0]
\email{qianfulan@hotmail.com}
\affiliation{
\institution{Key Laboratory of Intelligent Computing and Signal Processing, Ministry of Education, Anhui University}
\country{China}}

\renewcommand{\shortauthors}{Zheng and Wang, et al.}

\begin{abstract}
  Knowledge graph embeddings (KGE) have been validated as powerful methods for inferring missing links in knowledge graphs (KGs) that they typically map entities into Euclidean space and treat relations as transformations of entities. Recently, some Euclidean KGE methods have been enhanced to model semantic hierarchies commonly found in KGs, improving the performance of link prediction. To embed hierarchical data, hyperbolic space has emerged as a promising alternative to traditional Euclidean space, offering high fidelity and lower memory consumption. Unlike Euclidean, hyperbolic space provides countless curvatures to choose from. However, it is difficult for existing hyperbolic KGE methods to obtain the optimal curvature settings manually, thereby limiting their ability to effectively model semantic hierarchies. To address this limitation, we propose a novel KGE model called \textbf{Hyp}erbolic \textbf{H}ierarchical \textbf{KGE} (HypHKGE). This model introduces attention-based learnable curvatures for hyperbolic space, which helps preserve rich semantic hierarchies. Furthermore, to utilize the preserved hierarchies for inferring missing links, we define hyperbolic hierarchical transformations based on the theory of hyperbolic geometry, including both inter-level and intra-level modeling. Experiments demonstrate the effectiveness of the proposed HypHKGE model on the three benchmark datasets (WN18RR, FB15K-237, and YAGO3-10). The source code will be publicly released at https://github.com/wjzheng96/HypHKGE.
\end{abstract}




\maketitle

\section{Introduction}
Knowledge Graphs (KGs) provide descriptions of real-world entities and connections among them by structured representation, essentially a semantic network. KGs are stored as the collections of triples (\emph{head entity}, \emph{relation}, \emph{tail entity}) and denoted as $(h, r, t)$, each triple is a fact in the the real-world. KGs such as GoogleKG, Freebase~\cite{bollacker2008freebase}, YAGO~\cite{mahdisoltani2014yago3}, and NELL~\cite{carlson2010toward} have been applied in the fields of semantic search and information extraction. However, the coverage of these KGs is usually incomplete, and due to the huge amount of facts in the Big Data era, it is impractical to complete them manually. Therefore, many knowledge graph embeddings (KGE) methods have been proposed to automatically predict missing links in KGs, which is called the link prediction task.

\begin{figure}[htb] 
\centering 
\setlength{\belowcaptionskip}{0.1cm}
\setlength{\abovecaptionskip}{0.2cm}
\includegraphics[width=2.7in]{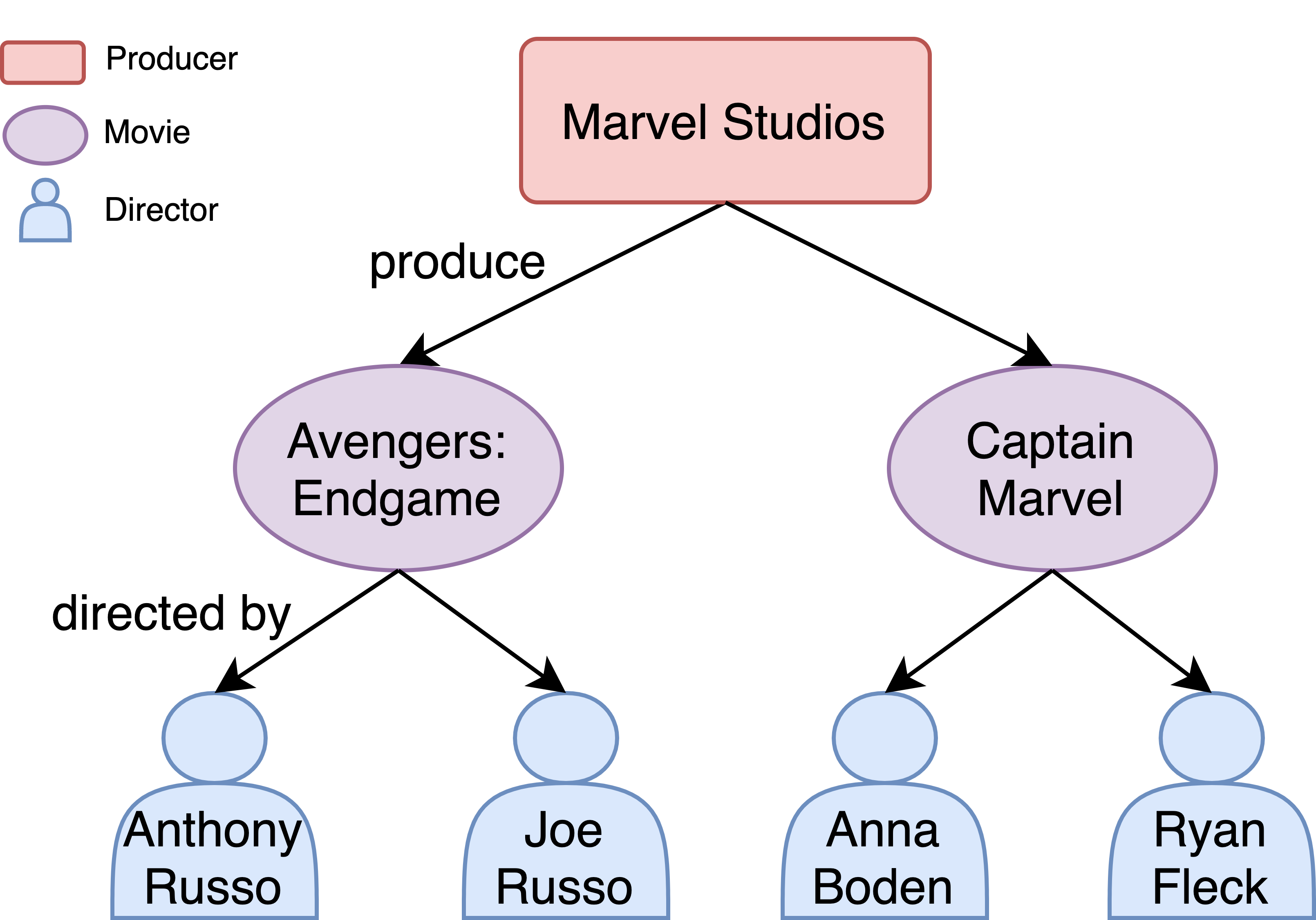} 
\caption{An example of semantic hierarchies in KGs where entities are linked through relations in the hierarchical structure.} 
\label{fig_0} 
\end{figure}

The goal of KGE is to embed entities and relations into a low-dimensional space while preserving rich potential information in KGs as much as possibly. For example, the earlier RotatE~\cite{sun2019rotate} takes into account the logical patterns of relations in the knowledge graph and modeled them using rotation. Recently, researchers have found that common semantic hierarchies (resemble tree-like structure, as shown in Fig~\ref{fig_0}) are implied in KGs~\cite{xie2016representation,zhang2018knowledge}, and some researches have tried to model semantic hierarchy in Euclidean space and got greate performance enhancement. For instance, HAKE \cite{zhang2020learning} maps entities into the polar coordinate system, which leverages modulus part and phase part to model the hierarchy. Lu et al.~\cite{lu2022dense} define the scaling operation in the 3-D Euclidean space to model the intrinsic semantic hierarchical structure of entities. Nevertheless, the expressiveness of their models over-relies on high embedding dimensionality that will cause large memory consumption.

For the above-mentioned problems about Euclidean KGE methods, hyperbolic geometry provides a great solution~\cite{Balazevic2019,kolyvakis2020hyperbolic,Chami2020}. Embedding KG with tree-like structures into hyperbolic space requires low dimensions and has relatively small distortion. Recently, some researchers try to propose hyperbolic KGE method. AttH~\cite{Chami2020} proposed to model the relation patterns of KGs in hyperbolic space by reflections and rotations. Unfortunately, it does not model rich semantic hierarchies. MuRP~\cite{Balazevic2019} treated each dimension of hyperbolic embeddings as a level of entities to model entities of different levels, but ignored the modeling of entities with the same level of hierarchy. HBE~\cite{pan2021hyperbolic} extended polar coordinate system to hyperbolic space for modeling hierarchies, but it overlooks the rich curvature modeling available in hyperbolic space, which limits the model's fitting ability.

In this paper, we seek to model semantic hierarchies of KGs in hyperbolic space and infer facts based on them. However, there are several challenges: (i) different from Euclidean space with fixed curvature of $0$, hyperbolic space can choose any negative curvature. It is necessary to choose appropriate curvature so that hyperbolic space has enough room to separate the child nodes of the original data, thus the hierarchical structure can be preserved with low distortion. Therefore, how to choose the suitable curvatures for various semantic hierarchies in KGs to learn low-distortion hyperbolic embeddings? (ii) in Euclidean space, we can use the modulus and the phase in the polar coordinate system to represent different levels and the same level in the hierarchy, respectively~\cite{zhang2020learning}. However, hyperbolic space does not have such concepts. So, how to represent semantic hierarchies in hyperbolic space? (iii) we need to consider relations as hierarchical transformations of entities when inferring facts of KGs based on semantic hierarchies, but Euclidean operations (vector inner product, scalar multiplication of matrix, etc.) are not applicable to hyperbolic space, so how to define hyperbolic operations to model the transformations?

To address the challenges above, we propose the \textbf{Hyp}erbolic \textbf{H}ierarchical \textbf{K}nowledge \textbf{G}raph \textbf{E}mbeddings (HypHKGE) approach, which models the semantic hierarchies of KGs in hyperbolic space. Specifically, we (i) consider various semantic hierarchical structures of KGs and thus propose attention-based learnable curvatures to learn low-distortion hyperbolic embeddings; (ii) utilize hyperbolic distance between hyperbolic embeddings and the origin to distinguish entities of different levels, which is inspired by the concept of depth in the tree. Also, for entities of the same level, we distinguish them by angle, which reflects the relative position of the entity at that level; (iii) propose hyperbolic hierarchical transformations (including hyperbolic inter-level and hyperbolic intra-level transformations), which capture the semantic hierarchies of KGs in hyperbolic space. In addition, hyperbolic hierarchical transformations also consider the modeling of relation patterns.

Our contributions are summarized as follows:
\vspace{-0.6em}
\begin{itemize}[leftmargin=0.15in]
\setlength\itemsep{0.0pt}
\item To obtain the optimal hyperbolic curvature for embedding hierarchical data, we propose an attention-based learnable curvature approach to preserve the different levels of knowledge graphs.

\item To enhance the modeling of semantic hierarchies, we define hyperbolic inter-level transformations and hyperbolic intra-level transformations, and theoretically prove that they are strict hyperbolic operations.  

\item Experiments on three benchmark datasets demonstrate the superiority of our proposed HypHKGE model over the state-of-the-art systems. 
\end{itemize}

\section{Related Work}
\label{sec2}

\subsection{Euclidean KGE}

Traditionally, KGs are usually embedded in Euclidean space. These models are mainly based on translation and rotation operations. TransE~\cite{bordes2013translating} is the first translational model, whose main idea is to use a relation-specific translation operation for the head entity. TransH~\cite{wang2014knowledge}, TransR~\cite{lin2015learning} and TransG~\cite{xiao2015transg} then utilize different strategies to improve this idea. In general, these translational models use a small number of parameters, but the drawback is that they cannot model relation patterns completely, such as symmetric, inversion. Thus, instead of translation, RotatE~\cite{sun2019rotate} first regards relations as rotations and successfully models the various relation patterns in the complex vector space. HAKE~\cite{zhang2020learning} considers the relation patterns while modeling the semantic hierarchy in KGs. 5$^\star$E~\cite{nayyeri20215} proposes a novel KGE model in projective geometry (supports inversion, reflection, translation, rotation, and homothety). These approaches achieve great results. But unfortunately, they require high embedding dimensions and incur significant memory overhead.

\subsection{Hyperbolic KGE}
Recently, with the advantages of low-dimensional hyperbolic embeddings in representing hierarchical data, it is gradually used by researchers for KGE. MuRP~\cite{Balazevic2019} is the first hyperbolic KGE method, which utilizes the translation of head entities with relations to reduce the hyperbolic distance from tail entities. However, the modeling in MuRP only considers entities at different levels, and ignores entities at the same level. AttH~\cite{Chami2020} learns hyperbolic embeddings based on rotations and reflections, which focuses on the modeling of relation patterns. M2GNN~\cite{wang2021mixed} proposes Mixed-Curvature Multi-Relational Graph Neural Network to address the embedding heterogeneous multi-relational graphs with non-uniform structures. In contrast, our HypHKGE models the attention-based learnable curvatures by considering various semantic hierarchies of KGs, and defines hyperbolic inter-level and hyperbolic intra-level transformations can effectively translate entities at different levels as well as under the same level.

\section{Preliminaries}
\label{sec3}
\subsection{KGE for Link Prediction}
Given a KG $\mathcal{G}$, consisting of a set of facts $\{(h, r, t)\}$, where $h,t \in |E|$ (the set of entities in $\mathcal{G}$) and $r \in |R|$ (the set of relations in $\mathcal{G}$). The first step of KGE methods for link prediction is to encode entities $h,t$ and relations $r$ into low-dimensional vectors. And then, they leverage the scoring function to transform the vectors of facts into single dimension numbers, where the value of numbers represents the possibility of facts being true. Therefore, it is worth mentioning that the main difference among different KGE methods is how to define the scoring function. Generally, the function is established in Euclidean space, while we focus on hyperbolic space where the embedding distortion is lower and the required dimensionality is lower.

\subsection{Hyperbolic Geometry}
Since we regard hyperbolic space as the embedding space, background knowledge about the hyperbolic geometry will be provided in this part, more rigorous and in-depth expositions see~\cite{spivak1975comprehensive,boothby1986introduction,lee2013smooth}.

\begin{figure}[tb]
\centering 
\setlength{\belowcaptionskip}{0.1cm}
\setlength{\abovecaptionskip}{0.1cm}
\includegraphics[width=2.7in]{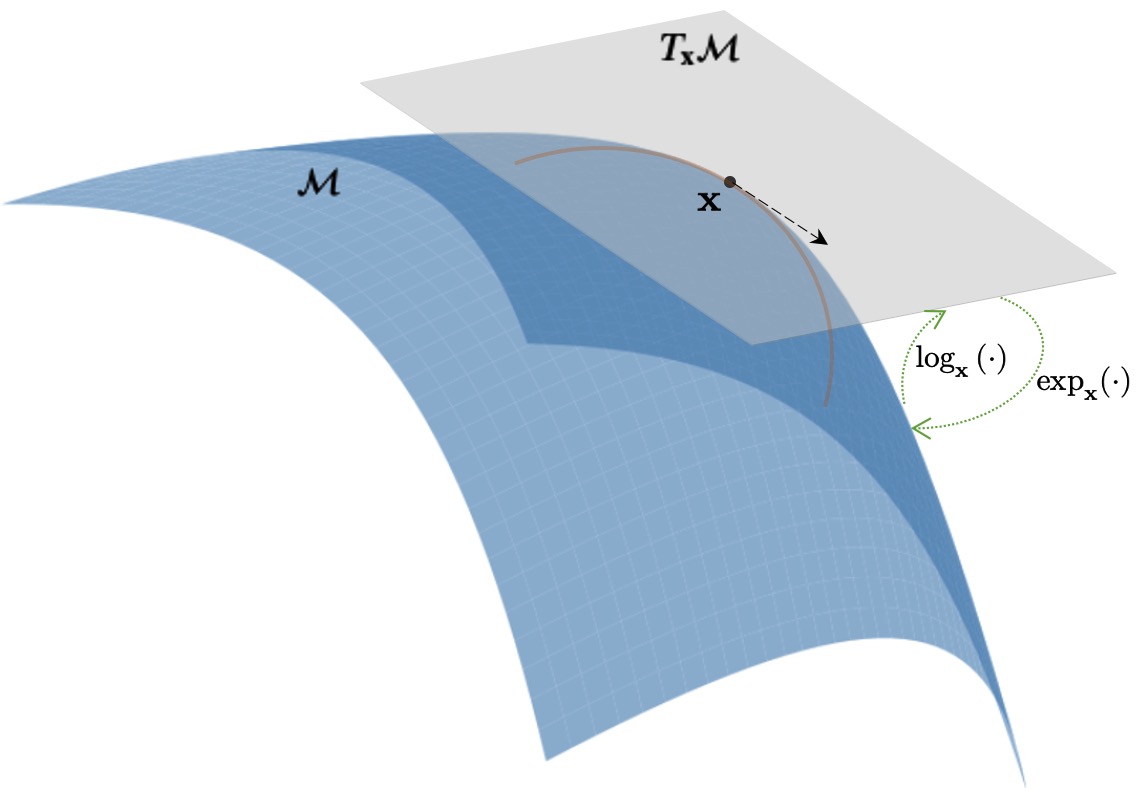} 
\caption{An illustration of the $log$ and the $exp$ maps. The $log$ map projects a point $\mathbf{x}\in \mathcal{M}$ to the tangent space $T_{\mathbf x}\mathcal{M}$ by moving a unit length along the geodesic, and the $exp$ map gives the reverse.} 
\label{fig_1} 
\end{figure}

\textbf{Hyperbolic Manifold}
In the theory of hyperbolic geometry, a hyperbolic manifold $\mathcal{M} $ is a space with a constant negative curvature $-c$ $(c>0)$, and the value of $c$ describes the degree of bending of hyperbolic manifold. For each point $\mathbf x\in \mathcal{M}$, one can define the tangent space $T_{\mathbf x}\mathcal{M}$ of $\mathcal{M} $. Intuitively, $T_{\mathbf x}\mathcal{M}$ is an Euclidean vector space which contains all possible directions that pass through $\mathbf x$ tangentially. Mapping between the manifold and the tangent space is performed by using $exponential$ and $logarithmic$ maps (see Fig. \ref{fig_1}). 

It is mentioned that the tangent space, $exponential$ and $logarithmic$ maps are useful to perform Euclidean operations undefined in hyperbolic space \cite{chami2019hyperbolic}. They will appear in Section~\ref{sec4}. 

\textbf{Poincar\'e Ball}
Hyperbolic space has some isometric models and we leverage the Poincar\'e ball model in this work~\cite{anderson2006hyperbolic,nickel2017poincare}. The Poincar\'e ball with a constant negative curvature corresponds to the hyperbolic manifold $(\mathbb{B}^{d,c},(g_{\mathbf x})_{\mathbf x})$, where $\mathbb{B}^{d,c}=\{\mathbf x\in \mathbb{R}^{d}\colon \| \mathbf x\|^{2} <\frac{1}{c} \}$, and $\| \cdot \| $ denotes the $L_2$ norm. 

Now, given $\mathbf{x},\mathbf{y} \in \mathbb{B}^{d,c} $, $\mathbf{v} \in T_{\mathbf o}\mathbb{B}^{d,c}$ (the tangent space of the origin), the $log$ map $\log_{\mathbf o}^{c}(\mathbf x):\mathbb{B}^{d,c} \rightarrow T_{\mathbf o}\mathbb{B}^{d,c}$ and the $exp$ map $\exp_{\mathbf o}^{c}(\mathbf v):T_{\mathbf o}\mathbb{B}^{d,c}  \rightarrow \mathbb{B}^{d,c}$ can be formulated as follows:

\begin{equation}
\label{eq1}
	\log^{c}_{\mathbf{0}}(\mathbf{x})  =  \frac{\mathrm{arctanh}( \sqrt{c} \| \mathbf{x}\| )}{\sqrt{c} \| \mathbf{x}\| }\mathbf{x}
\end{equation}

\begin{equation}
\label{eq2}
	\exp^{c}_{\mathbf{0}}( \mathbf{v})  = \frac{\tanh( \sqrt{c} \| \mathbf{v}\| )}{\sqrt{c} \| \mathbf{v}\| }\mathbf{v} 
\end{equation}
Also, the hyperbolic distance of $\mathbf{x}$ and $\mathbf{y}$ on $\mathbb{B}^{d,c}$ is defined as:
\begin{equation}
\label{eq3}
	\ d^{c}_{\mathbb{B}}( \mathbf{x},\mathbf{y})  =\frac{2}{\sqrt{c} } \mathrm{arctanh}( \sqrt{c} \| -\mathbf{x}\oplus^{c} \mathbf{y}\| )  
\end{equation}
where $\oplus^{c}$ denotes M\"{o}bius addition~\cite{ganea2018hyperbolic}:
\begin{equation}
\label{eq4}
	\ \mathbf{x}\oplus^{c} \mathbf{y}=\frac{( 1+2c\left<\mathbf{x},\mathbf{y}\right>  +c\| \mathbf{y}\|^{2}) \mathbf{x}+( 1-c\| \mathbf{x}\|^{2} )  \mathbf{y}}{1+2c\left< \mathbf{x},\mathbf{y}\right>  +c^{2}\| \mathbf{x}\|^{2}\|\mathbf{y}\|^{2} } 
\end{equation}

\subsection{Hierarchies and Curvatures}
\label{sec3.3}

Hyperbolic space can embed hierarchical data with low distortion. However, it holds only if we choose appropriate curvature. For a complete N-ary tree, a curvature is suitable only when the curvature of hyperbolic space and the degree of the tree satisfy a special condition~\cite{krioukov2010hyperbolic}. But, for KGs with multi-relational complex structures, there is no exact correspondence between curvature and graph structure.

Therefore, in this work, we propose an attention-based learnable curvature, and leverage the powerful fitting capabilities of deep learning to adaptively learn appropriate curvature, and thus embed hierarchical data with low distortion.

\section{Methodology}
\label{sec4}
In this section, we will present the implementation details of the proposed HypHKGE, which models the semantic hierarchies of KGs, even if it is restricted to low-dimensional hyperbolic space. To be specific, HypHKGE, 1) designs learnable curvatures that consider semantic hierarchies to obtain low distortion hyperbolic KGE; 2) describes how to represent the semantic hierarchies in hyperbolic embeddings, and then, we treat relations as transformations for hierarchies of entities and present hyperbolic hierarchical transformations; 3) presents our scoring function.

\subsection{KGE in Hyperbolic Space}
\label{sec4.1}
We firstly introduce the method of obtaining hyperbolic KGE indirectly, which is a prerequisite for the using of learnable curvatures.

In hyperbolic space, due to the existence of curvature, traditional methods of directly obtaining hyperbolic embeddings often need to specify the curvature first. However, under different semantic hierarchies, they can only learn hyperbolic embeddings with the fixed curvature, so that only lower-quality embeddings can be obtained.

To solve the questions raised above, we first model KGE in the tangent space of the hyperbolic origin, and then consider different hierarchical structures to learn the appropriate curvatures $c_{h,r}$. Next, we acquire initialized hyperbolic KGE through mapping it back to hyperbolic space.

This indirect method has been proven to be able to learn higher-quality hyperbolic KGE~\cite{sonthalia2020tree,Chami2020}.

\textbf{Learnable Curvatures} According to Section~\ref{sec3.3}, we know the importance of modeling the corresponding curvature for each hierarchical structure. Therefore, in this part, we consider different hierarchical structures to model curvatures.

\begin{figure}
\centering 
\setlength{\belowcaptionskip}{0.1cm}
\setlength{\abovecaptionskip}{0.1cm}
\includegraphics[width=3.5in]{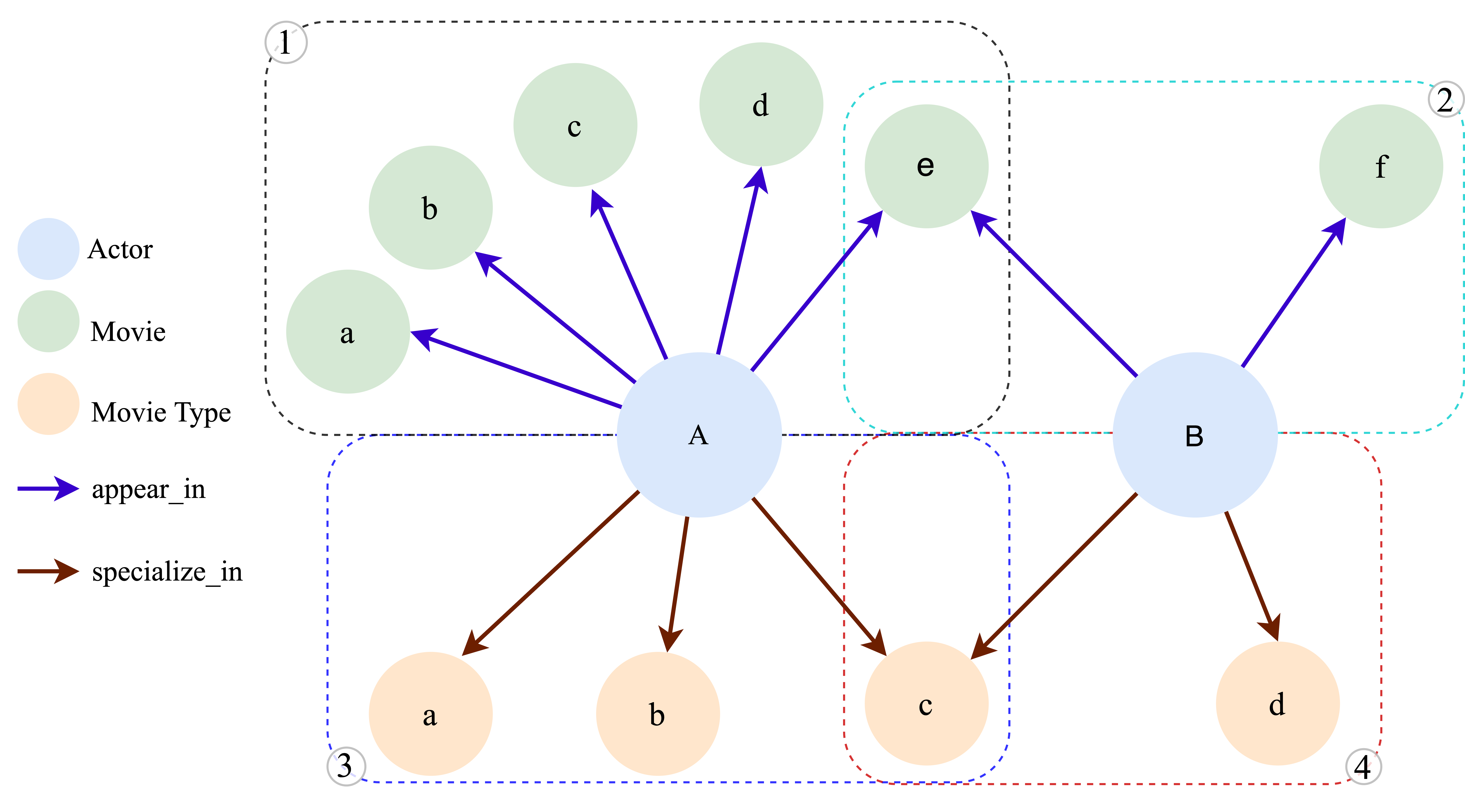} 
\caption{Spatial curvature modeled by different approaches from the perspective of semantic hierarchy. Assume that \textcircled{1}, \textcircled{2}, \textcircled{3}, and \textcircled{4} are the four hierarchical structures existing in KGs. MuRP~\cite{Balazevic2019} embeds the full graph in hyperbolic space with the fixed curvature, i.e., \textcircled{1}, \textcircled{2}, \textcircled{3}, and \textcircled{4} use the same curvature. AttH~\cite{Chami2020} learns a curvature for each relation, i.e., \textcircled{1} and \textcircled{2} share a curvature, \textcircled{3} and \textcircled{4} share another curvature. In contrast, our approach considers the head entity in addition to the relations, i.e., \textcircled{1}, \textcircled{2}, \textcircled{3}, and \textcircled{4} use unique curvature, respectively.}
\label{fig_3} 	
\end{figure}

First, entities in KGs have different hierarchical structures under different relations. For instance, the relations ``\emph{appear\_in}" and ``\emph{specialize\_in}" shown in Fig.~\ref{fig_3} correspond to different hierarchies. Therefore, AttH tried to model relation-specific curvatures and obtained the better embeddings than MuRP that sets a fixed curvature. Unfortunately, for modeling curvatures, AttH only considers the relations and ignores the influence of head entities. It can also be seen from Fig.~\ref{fig_3} that for the structure under various relations, we can further divide them based on the difference of head entities. This is because the hierarchical structure between different head entities can also vary under certain relations, as reflected by the fact that actors with different debut times have different numbers of movies in which they have appeared. Also, the magnitude of this difference varies for different relations. As shown in Fig.~\ref{fig_3}, the structure of the relation ``\emph{specialize\_in}" varies less as the ``\emph{Actor}" changes, while the structure of the relation ``\emph{appear\_in}" varies more as the ``\emph{Actor}" changes, reflecting the fact that although ``\emph{Actor A}" and ``\emph{Actor B}" have different debut times, the difference in the number of movie types they ``\emph{specialize\_in}" is not significant. However, for the number of movies they ``\emph{appear\_in}", the difference is large. 

Here, we construct learnable curvatures by considering the effects of relations and head entities together. In other words, we believe that for each class of triples ($h$, $r$, $\cdot$) (i.e., all triples with the specified head entity $h$ and relation $r$), the curvatures information is implicit in the attribute of $h$ and $r$. Therefore, we propose the modeling of curvatures based on attention mechanism, which takes into account all the above mentioned hierarchical structures. To be specific, let $\mathbf{h^E}$ and $\mathbf{r^E}$ denote head entity embeddings and relation embeddings in the tangent space of the origin, respectively. First, we define an attention vector $\mathbf{a} $ to fuse them, and thus obtain a fusion vector $\mathbf{v_{h,r}}$ which contains head entity and relation information:
\begin{equation}
	\ \alpha_{h} ,\alpha_{r} =\textrm{Softmax}( \mathbf{a}^{T}\mathbf{h^{E}},\mathbf{a}^{T}\mathbf{r^{E}})  
\end{equation}
\begin{equation}
	\ \mathbf{v_{h,r}} = \alpha_{h}\mathbf{h^E} + \alpha_{r}\mathbf{r^E} 
\end{equation}
where $\mathbf{a},\mathbf{v_{h,r}} \in \mathbb{R}^{d}$, $\alpha_{h}$, $\alpha_{r}$ is the attention value based on head entity and relation, respectively. When $\alpha_{h}\gg \alpha_{r} $, the modeled curvature is strongly influenced by the attributes of $h$ (i.e., the structure under this relation is strongly influenced by the head entity); when $\alpha_{h}\ll  \alpha_{r} $, the learned curvature is affected by the head entity attributes to a small extent.

Finally, we use a globally shared vector $\mathbf{p}\in \mathbb{R}^d$ and the softplus function to obtain the final curvature:
\begin{equation}
\label{eq7}
	\ c_{h,r} = \textrm{Softplus}(\mathbf{p}^{T}\mathbf{v_{h,r}})
\end{equation}

In summary, by defining attention-based learnable curvatures and trying to learn different hierarchical structures in KGs, we obtain initialized KGE of head entity $h$ and tail entity $t$ in hyperbolic space:
\begin{equation}
\label{eq8}
	\ \mathbf{h^H} = \exp_{\mathbf{0}}^{c_{h,r}}(\mathbf{h^E})
\end{equation}
\begin{equation}
\label{eq9}
	\ \mathbf{t^H} = \exp_{\mathbf{0}}^{c_{h,r}}(\mathbf{t^E})
\end{equation}

\subsection{Hyperbolic Hierarchical Transformations}
In this part, we first briefly explain how to describe the semantic hierarchy of KGs in hyperbolic space, and then introduce the proposed hyperbolic hierarchical transformations, which include hyperbolic inter-level transformations and hyperbolic intra-level transformations.

\subsubsection{Semantic Hierarchy in Hyperbolic Embeddings}
\cite{zhang2020learning} proposes that a high-quality semantic hierarchical representation should be able to distinguish not only entities of different levels of the hierarchy (like ``\emph{walk}" and ``\emph{run}"), but also entities of same level of the hierarchy (like ``\emph{rose}" and ``\emph{peony}"). Inspired by this, on the Poincar\'e disk, entities at different levels are distinguished by the hyperbolic distance of the entity and the origin of the Poincar\'e disk. Also, for entities with the same level,~\cite{nickel2017poincare} introduces that Poincar\'e disk is a conformal model of hyperbolic space and the angle between adjacent vectors is the same as that in Euclidean space. Therefore, we use the angle between the entity and the origin to distinguish them in hyperbolic space. From this perspective, semantic hierarchy is implicit in the hyperbolic embeddings, which distinguish not only entities of different levels but also entities of the same level.

The descriptions of semantic hierarchy mentioned above are performed in a two-dimensional Poincar\'e disk. For complex hyperbolic embeddings in $d$-dimensions ($d>2$ and is a multiple of 2), there is no unified way to describe them. In this work, we consider the $d$-dimensional hyperbolic embeddings as a composite of several two-dimensional hyperbolic embeddings, further to explain semantic hierarchies under the $d$-dimensional hyperbolic embeddings.

It is worth noting that the semantic hierarchy is an implicit property of the hyperbolic embeddings, and there is no direct metric rule based on the semantic hierarchy in hyperbolic space. In this paper, we transform semantic hierarchies by transforming the hyperbolic embeddings of entities.

\subsubsection{Hyperbolic Inter-Level Transformations} when embedding a KG with tree-like hierarchical structures into Poincar\'e disk, for any entity $h$, we denote the level of $h$ as $\hslash^{c}_{\mathbb{B}}(h)$. Formally, 
\begin{equation}
\label{eq4.1}
	\ \hslash^{c}_{\mathbb{B}}( h )  =d^{c}_{\mathbb{B}}( \mathbf{h^{H}},\mathbf{0})  
\end{equation}
where $\mathbf{h^{H}}\in \mathbb{B}^{2,c}$ represents the hyperbolic embeddings of $h$, and $d^{c}_{\mathbb{B}}( \mathbf{h^{H}},\mathbf{0})$ represents the hyperbolic distance of $h$ and the origin on the Poincar\'e disk.

For a real triple $(h,r,t)$, the level of $h$ and $t$ may be different, e.g., in (\emph{professor}, \emph{tenure}, \emph{university}). The effect of inter-level transformations is to make the transformed level of $h$ as similar as possible to the level of $t$. 

Therefore, let $h_e$ be the entity after the inter-level transformations, $\mathbf{h}_{e}^{\mathbf H}$ denotes the hyperbolic embeddings of $h_e$. The transformations that we required should satisfied the following conditions: 1) $\mathbf{h}^{\mathbf H}_e\in \mathbb{B}^{2,c}$ and 2) $\hslash^{c}_{\mathbb{B}}(h_e)=k\hslash^{c}_{\mathbb{B}}(h)$, where condition 1 indicates that we want $\mathbf{h}^{\mathbf H}_e$ not to be detached from the surface of the Poincar\'e disk with curvature $c$, condition 2 indicates the level of $h_e$ that we want is $k$ times of $h$ in hyperbolic space (note: there is no vector multiplication operation in hyperbolic space~\cite{spivak1975comprehensive,andrews2010ricci}).

To ensure that condition $1$ holds, we just map $h$ to the tangent space to accomplish operation, and then map it back to the hyperbolic space. But the question is how to define the operation on the tangent space to make condition $2$ hold? Here, We have explored this and the result is as follows: 

$\mathbf{Proposition}$ $\mathbf{1}$: While keeping the direction of the embedding vector constant, the level of $h$ in hyperbolic space becomes $k$ times of the original one, which is the sufficient and necessary condition for the level of $h$ in the tangent space that becomes $k$ times of the original one. That is, it is proved that:
\begin{equation}
	\ \hslash^{c}_{\mathbb{B}}(h_{e})  =k\hslash^{c}_{\mathbb{B}}(h)  \Leftrightarrow \hslash_{\mathbb{E}} ( h_{e})  =k\hslash_{\mathbb{E}}(h)  
\end{equation}
where $\hslash_{\mathbb{E}}(\cdot)$ denotes the level of arbitrary entity in the tangent space of the origin. For detailed proof, see APPENDIX ~\ref{app1}. 

From the above, the purpose of getting $\mathbf{h}_{e}^{\mathbf H}$ can be achieved with the help of tangent space.

First, in tangent space, to make the length of a vector $k$ times of its original length while keeping the direction unchanged, it is self-evident that we just need to make both components ($x$-axis component vector and $y$-axis component vector) $\frac{k}{\sqrt{2} }$ times of their original length, i.e., $\mathbf{h}^{\mathbf E}_e=\frac{k}{\sqrt{2}}\mathbf{h^E}$, where $\mathbf{h^E}$ and $\mathbf{h}^{\mathbf E}_e$ denote the embedding of $h$ in tangent space before and after transformations, respectively. Therefore, the representation of $h$ after the inter-level transformations in tangent space:
\begin{equation}
	\ \mathbf{h}^{\mathbf E}_e=\tau_r \mathbf{h^{E}} 
\end{equation}
where $\mathbf{h}^{\mathbf E}_e\in \mathbb{R}^{2}$, the two-dimensional inter-level transformations factor $\tau_r = \left[ \begin{matrix}k^{\prime }&0\\ 0&k^{\prime }\end{matrix} \right] $, $k^{\prime }=\frac{k}{\sqrt{2}}$. 

In this paper, we explain the higher-dimensional hyperbolic embeddings in terms of the composite of several two-dimensional Poincar\'e embeddings. Naturally, in hyperbolic space, for the $d$-dimensional hyperbolic embeddings, hyperbolic inter-level transformations be defined:
\begin{equation}
\label{inter-trans}
	\ \mathbf{h}^{\mathbf H}_e = \exp_{\mathbf{0}}^{c}(\mathbf{h}^{\mathbf E}_e) = \exp_{\mathbf{0}}^{c}(\daleth_r\mathbf{h^E})
\end{equation}
where $\mathbf{h}^{\mathbf H}_e\in \mathbb{B}^{d,c}$ be the hyperbolic embeddings after transformations, the $d$-dimensional inter-level transformations factor $\daleth_r=\text{diag}( \tau_{r,1} ,...,\tau_{r,\frac{d}{2} })$, $\tau_{r,{i}} = \left[ \begin{matrix}k_{i}&0\\ 0&k_{i}\end{matrix} \right]$, $i\in(1,2,...,\frac{d}{2})$.

\subsubsection{Hyperbolic Intra-level Transformations}
\label{sec4.2.2.1}
There are many entities at the same level, so they need to be further distinguished by using intra-level transformations. In this part, we model hyperbolic intra-level transformations by using Givens rotations, which not only can distinguish entities at the same level, but also have been demonstrated to infer the relation patterns of KGs~\cite{xu2019relation,Chami2020}.

Relations in KGs usually exhibit various logical patterns, such as symmetric/antisymmetric, inverse and composition, etc. The relation patterns portray the special properties of the relations in KGs. For instance, the relation ``\emph{is\_friend}" exhibits the symmetric pattern, and if the (\emph{A}, \emph{is\_friend}, \emph{B}) triplet is true, then the (\emph{B}, \emph{is\_friend}, \emph{A}) triplet also must be true.

To be specific, the Givens rotations defined in two-dimensional Euclidean space is :
\begin{equation}
	\ \mathbf{G}_y = \text{Rot}(\Theta_r)\mathbf{G}_x
\end{equation}
where $\Theta_r$ denotes the angle of rotation, $\mathbf{G}_x$ represents the vector before transformations and $\mathbf{G}_y$ represents the vector after transformations. Also, the rotations factor Rot$(\Theta_r)$ is defined as:
\begin{equation}
	\ \text{Rot}( \Theta_r )  =\left[ \begin{matrix}\cos( \Theta_r )  &-\sin( \Theta_r )  \\ \sin( \Theta_r )  &\cos( \Theta_r )  \end{matrix} \right]  
\end{equation}
Then, the $d$-dimensional intra-level transformations factor is defined as $ \Theta_r =\text{diag}( \text{Rot}( \Theta_{r,1}  )  ,...,\text{Rot}( \Theta_{r,\frac{d}{2}} ) ) $. 

We conclude that the direct use of the $\Theta_r$ in hyperbolic space does not detach the hyperbolic embeddings out of the hyperbolic manifold. Formally,

$\mathbf{Proposition}$ $\mathbf{2}$: For entity $h$, given the hyperbolic embeddings $\mathbf{h^H}$ and the tangent embeddings $\mathbf{h^E}$, using the intra-level transformations factor $\Theta_r$ in hyperbolic space have the same effect as using in tangent space. That is, given $\mathbf{h^H}\in \mathbb{B}^{d, c}$ and $ \mathbf{h^E} = \log_{\mathbf{0}}^{c}(\mathbf{h^H})$, we conclude that:
\begin{equation}
	\ \exp_{\mathbf{0}}^{c}(\Theta_r\mathbf{h^E}) =\Theta_r\mathbf{h^H}
\end{equation}
For detailed proof, see APPENDIX~\ref{app2}.

According to Proposition 2, the representation of the head entity $h$ after the $d$-dimensional hyperbolic intra-level transformations:
\begin{equation}
\label{intra-trans}
	\ \mathbf{h}^{\mathbf{H}}_{rot} = \Theta_r\mathbf{h^H}
\end{equation} 
where $\mathbf{h}^{\mathbf{H}},\mathbf{h}^{\mathbf{H}}_{rot} \in \mathbb{B}^{d,c}$ and $\mathbf{h}^{\mathbf{H}}$ are the hyperbolic embeddings of $h$ before intra-level transformations.

\subsection{Scoring Function}
\label{sec4.3}
In this part, we describe how to construct scoring function. 

For triple ($h$, $r$, $t$), we firstly compute the attention-based learnable curvatures $c_{h,r}$ to obtain the initial hyperbolic embeddings of head entity $h$ (Eq.~\ref{eq8}) and tail entity $t$ (Eq.~\ref{eq9}).

Next, we perform hyperbolic hierarchical transformations for $h$ on the $\mathbb{B}^{d,c_{h,r}}$ based on Eq.~\ref{inter-trans} and Eq.~\ref{intra-trans}:

%
\begin{equation}
	\ \mathbf{h}_{e,rot}^{\mathbf H} =\Theta_r\exp_{\mathbf{0}}^{c_{h,r}}(\daleth_r\mathbf{h^E})
\end{equation}
where $\mathbf{h}_{e,rot}^{\mathbf H}$ denotes the hyperbolic embeddings after inter-level transformations and intra-level transformations, $\daleth_r$ and $\Theta_r$ are corresponding $d$-dimensional transformation factors.

Finally, we perform a translation on the head entity $h_{e,rot}$, and further perform a similarity measure with the tail entity $t$ by the hyperbolic distance (Eq.~\ref{eq3}). The final scoring function is defined as:
\begin{equation}
\label{eq_score}
\ s(h,r,t)  = -d^{c_{h,r}}_{\mathbb B}( \mathbf{h}^{\mathbf H}_{e,rot}\oplus^{c_{h,r}} {\mathbf \varepsilon}^{\mathbf H}_{r},\mathbf{t}^{\mathbf H})^2 +b_{h}+b_{t} 
\end{equation}
where ${\mathbf \varepsilon}^{\mathbf H}_{r}\in \mathbb{B}^{d,c_{h,r}}$ is a hyperbolic translation vector. Also, $b_h$ and $b_t$ are entity biases which act as margins in the scoring function~\cite{tifrea2019poincar}.

\section{Experiments and Analysis}
\label{sec5}
\subsection{Experimental setup}

\subsubsection{Datasets}
Three common KGs (WN18RR~\cite{bordes2013translating,dettmers2018convolutional}, FB15k-237~\cite{bordes2013translating,toutanova2015observed} and YAGO3-10~\cite{mahdisoltani2014yago3}) are used as benchmark datasets in our experiments, and further evaluate the performance of HypHKGE for the link prediction task. Among them, WN18RR is used to describe the KG of relations between words with a natural hierarchical structure~\cite{Chami2020,wang2021mixed}; FB15K-237 is constructed by knowledge base relation triples and textual mentions of a web-scale corpus; Each entity in YAGO3-10 has at least 10 relations and most of the relations are descriptions of people (e.g., gender, age). In addition, to cope with the test data leakage problem, the WN18RR and the FB15K-237 do not have inverse relations~\cite{tang2020orthogonal}. The statistics of three KGs are summarized in Table~\ref{tab_data}. 

\begin{table}[htb]
\centering
\begin{threeparttable}
\setlength{\belowcaptionskip}{0.1cm}
\setlength{\abovecaptionskip}{0.2cm}
\setlength{\tabcolsep}{14pt}
\begin{tabular}{cccccc}
\toprule
Dataset   & $|E|$     & $|R|$  & \#TR      & \#VA   & \#TE   \\ \midrule
WN18RR    & 40,943  & 11  & 86,835    & 33,034 & 3,134  \\
FB15K-237 & 14,541  & 237 & 272,115   & 17,535 & 20,466 \\
YAGO3-10  & 123,182 & 37  & 1,079,040 & 5,000  & 5,000  \\ \bottomrule
\end{tabular}
\caption{Dataset properties. $|E|$ and $|R|$ respectively denote the number of entities and the number of relations. \#TR, \#VA, and \#TE represent the number of train set, validation set, and test set, respectively.}
\label{tab_data}
\end{threeparttable}
\end{table}

\subsubsection{Baselines}
\textbf{ComplEx}~\cite{trouillon2016complex} models Euclidean embedding with a diagonal relational matrix in a complex space. \textbf{RotatE}~\cite{sun2019rotate} proposes a rotational method to model logical patterns of relations based on Euclidean complex space. \textbf{HAKE}~\cite{zhang2020learning} maps entities into polar coordinate system (Euclidean) with modulus part and phase part. \textbf{MuRP} \& \textbf{MuRE}~\cite{Balazevic2019} learn relation-specific parameters to transform entities embeddings. Note that MuRE is a Euclidean extension version of MuRP, which sets the curvature to zero. \textbf{AttH} \& \textbf{AttE}~\cite{Chami2020} use rotation, reflection, and translation to model relation patterns. Note that AttE is a Euclidean extension version of AttH, which sets the curvature to zero. \textbf{5$\star$E}~\cite{nayyeri20215} models a relation as five transformation functions (includes inversion, reflection, translation, rotation, and homothety). \textbf{MuRMP}~\cite{wang2021mixed} is a generalization of MuRP, which embeds entities to a mixed-curvature space. Similarly, we design the Euclidean extension version of HypHKGE as a baseline, called EucHKGE, which sets the curvature to zero.

\subsubsection{Training Protocol and Evaluation Metrics}
To train the model, we use the cross-entropy loss function with uniform negative sampling~\cite{Chami2020}:
\begin{equation}
	\ \mathcal{L}=\sum_{h^{\prime },t^{\prime }\sim \mathcal{U}( \mathcal{V})  } \log( 1+\exp( y \cdot s( h^{\prime },r,t^{\prime } )  ) )  
\end{equation} 
\centerline{where $y=\begin{cases}-1&\text{if}\  h^{\prime }=h $ and $ t^{\prime }=t\\ 1&\text{otherwise}\end{cases}$}

For each triplet $(h, r, t)$, we replace head entity $h$ or tail entity $t$ with a candidate entity to construct the set of candidate triples $(?, r, t)$ or $(h, r, ?)$, and then rank them in descending order using the scoring function (Eq.~\ref{eq_score}). We use the ``filtered" setting protocol~\cite{bordes2013translating}, and all true triples in KG are filtered out, because predicting the low rank of these triples should not be punished. We find the optimal hyperparameters based on the results of the validation set through grid search. 
 
Moreover, we use two common evaluation metrics: mean reciprocal rank (MRR) and hits at K (H@K, K$\in{1, 3, 10}$). Higher MRR or higher Hits@10 means greater performance of the model.

\subsection{Results in Low Dimensions}
\label{sec5.2}
First, we show the experimental results of HypHKGE in low dimensions (d=32). Table~\ref{table_low_dim} shows performance comparison between HypHKGE and baselines.

Comparing to Euclidean embeddings SOTA approach, HypHKGE improves 4.1\%, 2.2\% and 14.2\% in MRR for the WN18RR, FB15K-237, and YAGO3-10, respectively. It is shown that all datasets are hierarchical and the hierarchical structure under the imposition of hyperbolic geometry learns better embeddings. In particular, the comparison with HAKE (modeling semantic hierarchies in polar coordinate system) shows that the hyperbolic embeddings can better model the semantic hierarchies of KGs in scenarios that require low embedding dimensionality. This is a great news for researchers stuck with high memory consumption.

Comparing hyperbolic embeddings SOTA approach, we have improved 1.5\%, 1.9\% and 7.6\% in MRR for the WN18RR, FB15K-237 and YAGO3-10, respectively. Compared with them, HypHKGE i) is more concerned with various hierarchical structures in KGs. To be specific, we define a method that models curvatures based on attention mechanism. Compared to MuRP setting for fixed curvature and AttH setting for relation-specific curvatures, we believe that the role played by head entities in the modeling of curvatures is equally indelible. Moreover, to further explore the influence on the quality of hyperbolic embeddings by different methods of curvatures modeling, we conduct experiments in Section~\ref{sec5.5}; ii) makes better use of the advantage of hyperbolic geometry. AttH focuses on modeling relation patterns in hyperbolic space, and fails to take great advantage of hyperbolic geometry to model hierarchies. Also, the hierarchy modeling of MuRP is incomplete, which only cares for the different levels of hierarchies. In contrast, we model different levels and same level of hierarchies and capture relation patterns by utilizing hyperbolic geometry.

\begin{table}[tb]
\centering
\begin{threeparttable}
\setlength{\belowcaptionskip}{0.1cm}    
\setlength{\abovecaptionskip}{0.1cm}
\scalebox{0.78}{
\begin{tabular}{lccccccccccccc}
\toprule
\multirow{2}{*}{Space} & \multirow{2}{*}{Model} & \multicolumn{4}{c}{WN18RR}                                                                             & \multicolumn{4}{c}{FB15K-237}                                                                          & \multicolumn{4}{c}{YAGO3-10}                                                                           \\ \cmidrule(r){3-6}\cmidrule(r){7-10}\cmidrule(r){11-14}
\multicolumn{1}{c}{}                   &                        & \multicolumn{1}{l}{MRR} & \multicolumn{1}{l}{H@1} & \multicolumn{1}{l}{H@3} & \multicolumn{1}{l}{H@10} & \multicolumn{1}{l}{MRR} & \multicolumn{1}{l}{H@1} & \multicolumn{1}{l}{H@3} & \multicolumn{1}{l}{H@10} & \multicolumn{1}{l}{MRR} & \multicolumn{1}{l}{H@1} & \multicolumn{1}{l}{H@3} & \multicolumn{1}{l}{H@10} \\ \midrule 
\multirow{6}{*}{$\mathbb{R}^{d}$}                                      
                                      & ComplEx~\cite{trouillon2016complex}               &  .367                       & .326                         &  .397                       & .433                         &  .202                       & .116                        &  .219                       &  .384                        &  .198                      &  .116                       &  .223                       &  .359                        \\                                  
                                    & RotatE~\cite{sun2019rotate}                & .387                   & .330                    & .417                   & .491                    & .290                    & .208                   & .316                   & .458                    & .235                        &  .153                       &   .260                      &  .410                        \\                                
                                   & HAKE~\cite{zhang2020learning}              & .416                        & .389                        &  .427                       &  .467                        & .296                        &  .212                       &  .323                       &  .463                        &  .253                       &   .164                      &  .286                       &   .430                       \\ 
                                     & MuRE~\cite{Balazevic2019}                   & .458                   & .421             & .471                   & .525                    & .313                   & .226                   & .340                    & .489                    & .283                         & .187                        & .317                        &  .478                        \\
                                     	
                                     & AttE~\cite{Chami2020}                  & .456                   & .419                   & .471                   & .526                    & .311                   & .223                   & .339                   & .488                    & .374                        & .290                         & .410                        & .537                         \\
                                     & 5$^\star$E~\cite{nayyeri20215}                  & .449                   & .418                   & .462                   & .510                    & .323                   & .240                   & .355                   & .501                    & ---                      & ---                         & ---                        & ---                       \\

                                     & EucHKGE (Ours)                  & .462                        & .425                        &  .474                       & .534                         & .319                        & .228                        & .351                        & .499                         & .391                       & .308                        & .432                        & .550                         \\ \midrule

$\mathbb{B}^{d,1}$                                      & MuRP~\cite{Balazevic2019}                  & .465                   & .420                    & .484             & .544                    & .323                   & .235                   & .353                   & .501                    & .230                    & .150                    & .247                   & .392                    \\
$\mathbb{B}^{d,1}$                                      & MuRMP~\cite{wang2021mixed}                  & .470             & .426                   & .483            & .547              & .319             & .232             & .351             & .502              & .395             & .308             & .429             & .566              \\
$\mathbb{B}^{d,c_{r}}$                                      & AttH~\cite{Chami2020}                  & .466             & .419                   & .484            & .551              & .324             & .236             & .354             & .501              & .397             & .310             & .437             & .566               \\
$\mathbb{B}^{d,c_{h,r}}$                                      & HypHKGE (Ours)               & \textbf{.477}          & \textbf{.437}          & \textbf{.492}          & \textbf{.556}           & \textbf{.330}          & \textbf{.240}          & \textbf{.361}          & \textbf{.510}            & \textbf{.427}          & \textbf{.344}          & \textbf{.470}          & \textbf{.588}         \\ \bottomrule
\end{tabular}}
\caption{Link prediction results for low-dimensional embeddings ($d=32$) on WN18RR, FB15K-237, and YAGO3-10.The best results are in bold.}
\label{table_low_dim}
\end{threeparttable}
\end{table}

\subsection{Ablation Studies}
\label{sec5.3}
In this part, we perform ablation studies on the hyperbolic inter-level transformations part, the hyperbolic inter-level transformations part and the learnable curvatures part of HypHKGE to explore the role they play in modeling semantic hierarchies of KGs. Table~\ref{table_abs} shows the performance of these parts on three datasets. 

We observe that the setting of learnable curvatures can improve the performance of HypHKGE on all metrics. In particular, on the YAGO3-10 dataset, adding the learnable curvatures part to the hyperbolic inter-level transformations, hyperbolic intra-level transformations, and the combination of them increase the MRR metrics by 6.9\%, 6.4\%, and 8.1\%, respectively. This illustrates the importance of modeling learnable curvatures by considering various semantic hierarchies. 

We also observe that the hyperbolic inter-level and hyperbolic intra-level transformations parts of HypHKGE reflect different results on different datasets, such as the result only based on hyperbolic inter-level transformations is higher than the result only based on hyperbolic intra-level transformations on the WN18RR dataset, but the show is opposite on the FB15K-237. These results illustrate that datasets have different dependencies on modeling different levels of hierarchy or the same level of hierarchy. However, our HypHKGE is remarkably superior to the hyperbolic inter-level transformations part and the hyperbolic intra-level transformations part on all datasets, which demonstrates the importance of combining these parts for the modeling of semantic hierarchies. 

\begin{table*}[t]
\centering
\small
\begin{threeparttable}
\setlength{\belowcaptionskip}{0.1cm}
\setlength{\abovecaptionskip}{0.1cm}
\scalebox{0.9}{
\begin{tabular}{ccccccccccccccc}
\toprule
                      &                       &                       & \multicolumn{4}{c}{WN18RR}                                                                                                                                                        & \multicolumn{4}{c}{FB15K-237}                                                                                                                                                  & \multicolumn{4}{c}{YAGO3-10}                                                                           \\ \cmidrule(r){4-7} \cmidrule(r){8-11} \cmidrule(r){12-15}
\multicolumn{1}{c}{$\daleth_r$} & \multicolumn{1}{c}{$\Theta_r$} & \multicolumn{1}{c}{$c_{h,r}$} & \multicolumn{1}{l}{MRR}                    & \multicolumn{1}{l}{H@1}                    & \multicolumn{1}{l}{H@3}                    & \multicolumn{1}{l}{H@10}                   & \multicolumn{1}{l}{MRR}                   & \multicolumn{1}{l}{H@1}                   & \multicolumn{1}{l}{H@3}                    & \multicolumn{1}{l}{H@10}                  & \multicolumn{1}{l}{MRR} & \multicolumn{1}{l}{H@1} & \multicolumn{1}{l}{H@3} & \multicolumn{1}{l}{H@10} \\ \midrule
                  \checkmark        &                   &                       & .431                                      & .394                                      & .450                                      & .497                                      & .323                                     & .232                                     & .355                                     & .503                                     & .361                   & .269                   & .408                   & .526   \\

                      &        \checkmark               &                       & .463                                      & .415                                      & .486                                      & .554                                      & .310                                     & .219                                     & .344                                      & .492                                     & .360                   & .277                   & .398                   & .522 \\
                      \checkmark       &        \checkmark               &                       & .466                                     & .419                                      & .489                                      & .552                                      & .324                                     & .234                                     & .356                                      & .503                                     & .395                   & .313                   & .435                   & .553  
                                       \\    \hline  
                     \checkmark         &                      &  \checkmark                     & .450                                      & .414                                      & .467                                      & .519                                     & .325                                     & .236                                     & .357                                      & .505                                     & .386                   & .295                   & .436                   & .557                    \\ 
         &      \checkmark                 &  \checkmark                                                &    .469                                   &   .427                                    &   .486                                    &   .551                                  &    .316                                 &  .225                                     &   .348                                  &  .499                &   .383                  & .291          &   .429     &  .552            \\      
             \checkmark       &         \checkmark               &     \checkmark                   & \textbf{.477} &  \textbf{.437} &  \textbf{.492} &  \textbf{.556} &  \textbf{.330} &  \textbf{.240} &  \textbf{.361} &  \textbf{.510} & \textbf{.427}                   & \textbf{.344}                   & \textbf{.470}                   & \textbf{.588}   \\                
\bottomrule
\end{tabular}}
\caption{Ablation results on WN18RR, FB15k-237 and YAGO3-10 datasets. The symbols $\daleth_r$, $\Theta_r$ and $c_{h,r}$ represent hyperbolic inter-level transformations part, hyperbolic intra-level transformations part and learnable curvatures part of HypHKGE, respectively.}
\label{table_abs}
\end{threeparttable}
\end{table*}

\subsection{HypHKGE vs. EucHKGE}
\label{sec5.4}
In this part, we compare HypHKGE and EucHKGE to illustrate the benefit of hyperbolic embeddings.

\begin{figure}[htb]
\centering
\setlength{\belowcaptionskip}{0.1cm}
\setlength{\abovecaptionskip}{0.1cm}
\includegraphics[width=3.3in]{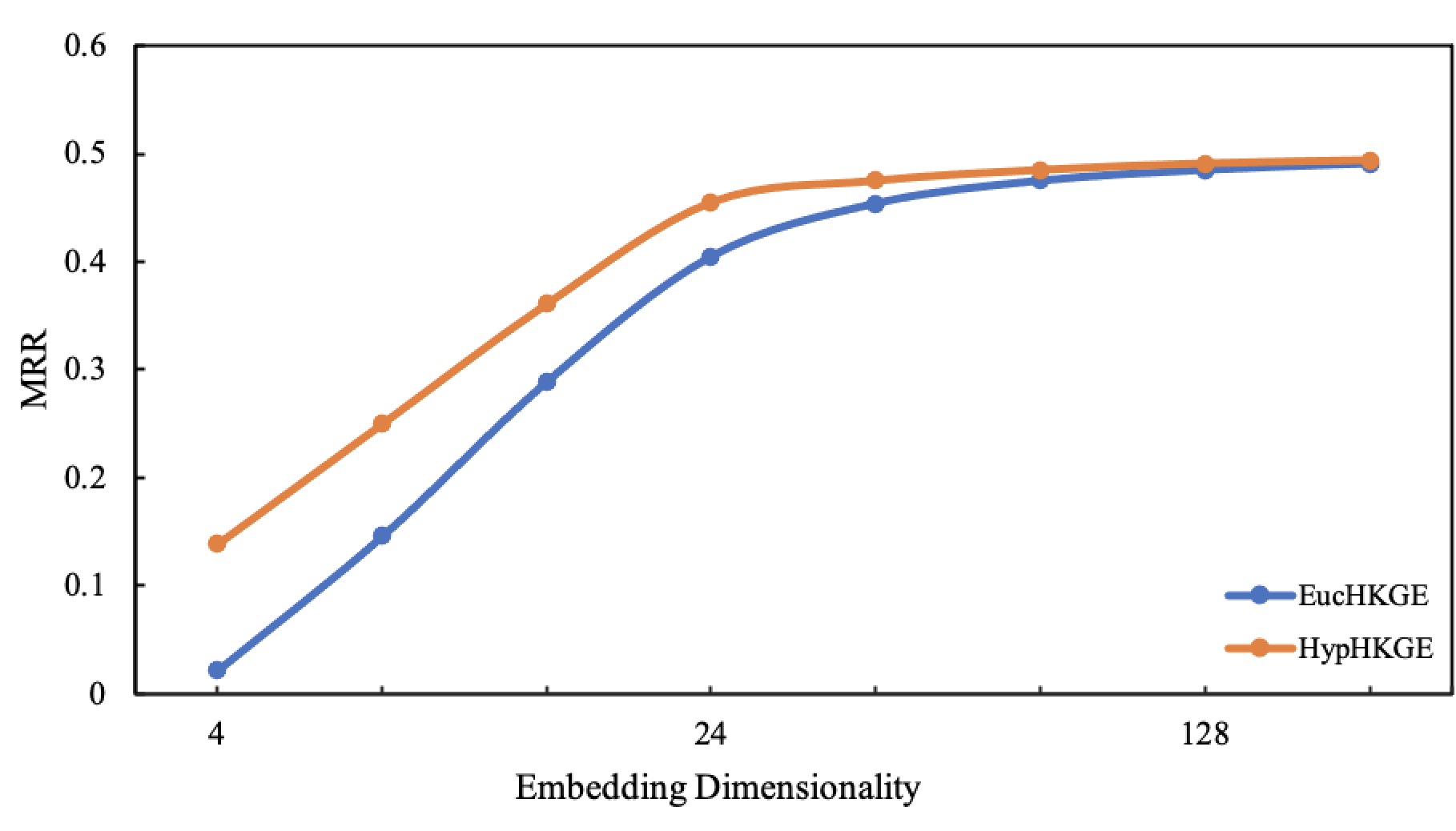} 
\caption{With the embedding dimension increases, comparison of MRR of HypHKGE and EucHKGE on WN18RR.} 
\label{fig_4} 
\end{figure}

\textbf{Effect of dimensionality} To demonstrate the advantage of low-dimensional embeddings in hyperbolic space, we compared the proposed HypHKGE and its Euclidean version (EucHKGE) on the WN18RR in different dimensions taken as $\left\{4,8,12,24,32,64,128,256\right\}$. As shown in Fig.~\ref{fig_4}, we observe that at lower embedding dimensions, the difference is greater for HypHKGE compared to EucHKGE. However, as the dimensionality rises to $64$ or higher dimensions, both the Euclidean space and the hyperbolic space have sufficient capacity to learn the semantic hierarchy, so the difference between them gradually shrinks.

\textbf{Performance per relation} To demonstrate the advantage of capturing hierarchical relations in low dimensional hyperbolic space, we compare HypHKGE and EucHKGE on different relations for the WN18RR (dim = 32). As shown in Table~\ref{table_relation}, relations are ranked by $\text{Khs}_{\mathcal{G}}$~\cite{Balazevic2019} and $\xi_{\mathcal G}$~\cite{Chami2020}, for relations such as ``verb\_group" and ``similar\_to" have little hierarchy, we can find that the performance of HypHKGE and EucHKGE is comparable. On the contrary, for relations like ``member\_meronym" and ``hypernym", the higher hierarchy, the greater difference between HypHKGE and EucHKGE. This also reflects that it is more profitable to learn hierarchical representation in hyperbolic space than in Euclidean space.

\begin{table}[htb]
\centering
\small
\begin{threeparttable}
\setlength{\belowcaptionskip}{0.1cm}
\setlength{\abovecaptionskip}{0.1cm}
\setlength{\tabcolsep}{10pt}
\scalebox{0.95}{
\begin{tabular}{lccccc}
\toprule
Relation                           & \multicolumn{1}{c}{$\text{Khs}_{\mathcal{G}}$}              & \multicolumn{1}{c}{$\xi_{\mathcal G}$}                                     & EucHKGE                               & HypHKGE                                                   &  $\maltese$     \\ \midrule
member\_meronym                 & 1.00    &  -2.90   &     .326    &    \textbf{.407}      &     24.9\%           \\
hypernym                      &    1.00                                  &   -2.46                                                        &   .226                              &        \textbf{.263}                                                      &    16.4\%           \\
has\_part                     &  1.00                                    & -1.43    & .288 & \textbf{.334} & 16.0\% \\
instance\_hypernym             & 1.00                                                           &       -0.82                                                    &  .492         &    \textbf{.520}   &   10.6\%   \\
member\_of\_domain\_region      &  1.00    &   -0.78     &   .385                                                        &   \textbf{.442}      &   14.8\%               \\
member\_of\_domain\_usage       &  1.00    &   -0.74            &    \textbf{.458}               &    .438                                                     &   -4.37\%                                                       \\
synset\_domain\_topic\_of      &  0.99  & -0.69     &   .386          &    \textbf{.425}            &    10.1\%                                                       \\  
also\_see      & 0.36          & -2.09      &  .634        &    \textbf{.643}      &   1.50\%                                   \\
derivationally\_related\_form  & 0.07 & -3.84            &   .959       &  \textbf{.964}        &      0.50\%                              \\
similar\_to                      &  0.07              &   -1.00                                                        &  \textbf{1.00}               &   \textbf{1.00}                                                        &   0.00\%  \\
verb\_group                     &   0.07                     &  -0.50                                     &  \textbf{.974}                                     &     \textbf{.974}                                &   0.00\%                                \\
\bottomrule
\end{tabular}}
\caption{Comparison of H@10 for WN18RR relations. Higher $\text{Khs}_{\mathcal{G}}$ and lower $\xi_{\mathcal G}$ means more hierarchical. $\maltese$ indicates the improvement rate.}
\label{table_relation}
\end{threeparttable}
\end{table}

\subsection{Attention-based Curvatures}
\label{sec5.5}

\subsubsection{Attention-based learnable curvatures on WN18RR}
We embed WN18RR on a two-dimensional Poincar\'e Ball, and then calculate the average curvature of the entity on different relations, as shown in Table~\ref{table_avg_degree}. It can be seen that the average curvature under different relations will be different. If setting the same curvature for all relations, there will undoubtedly be a large distortion. Furthermore, it can be observed that curvature and degree do not correspond precisely. This is mainly because the structure of knowledge graph is not only affected by the degree, but also by various complex structures such as multi-relations and loops between entities. Therefore, an accurate curvature cannot be set directly based on the magnitude of the degree.

\begin{table}[htb]
\centering
\small
\begin{threeparttable}
\setlength{\belowcaptionskip}{0.1cm}
\setlength{\abovecaptionskip}{0.1cm}
\setlength{\tabcolsep}{14pt}
\scalebox{0.95}{
\begin{tabular}{lcc}
\toprule
Relation        & Absolute Value of Curvature        & Average Out-degree  \\ \midrule
also\_see      &  0.23   &  1.84   \\
derivationally\_related\_form      &  1.21   &  1.84   \\
verb\_group & 0.96    & 2.43 \\
has\_part & 1.11 & 1.02 \\
hypernym &6.42  & 1.18 \\
instance\_hypernym &1.07  & 2.39 \\	
member\_meronym &3.68 & 8.09 \\
member\_of\_domain\_region & 1.31   & 25.16  \\
member\_of\_domain\_usage & 2.47 & 1.03 \\
similar\_to & 0.44 & 1.04 \\
synset\_domain\_topic\_of & 0.27 & 1.16 \\
\bottomrule
\end{tabular}}
\caption{The average curvature of the entity on different relations.}
\label{table_avg_degree}
\end{threeparttable}
\end{table}

\subsubsection{Effect of entity on the attention-based curvatures}
Similarly, we embed WN18RR into a two-dimensional Poincar\'e Ball and select four relations to explore the effect of entity information on curvature under a single relation.

\begin{figure}[t] 
\centering 
\setlength{\belowcaptionskip}{0.1cm}
\setlength{\abovecaptionskip}{0.1cm}
\subfigure{
\includegraphics[width=0.48\linewidth]{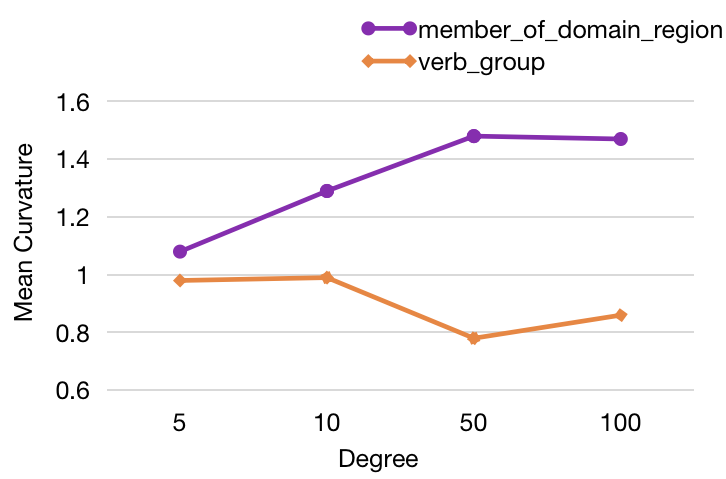}
}
\subfigure{
\includegraphics[width=0.48\linewidth]{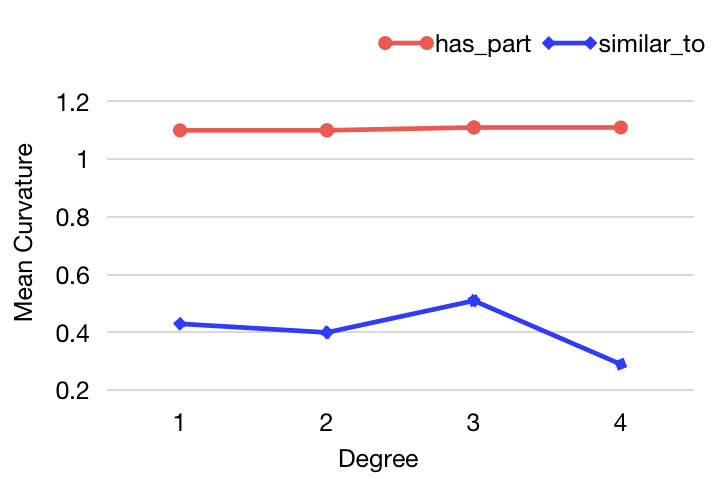}
}
\caption{Effect of entity information on curvature under a single relation.} 
\label{fig:degree_curva}  
\end{figure}

As shown in Fig.~\ref{fig:degree_curva}, among four kinds of relations, for the ``\emph{member\_of\_domain\_region}" and ``\emph{verb\_group}", distribution range of entity degree are wide, but the level of the former is strong, while the level of the latter is weak. ``\emph{has\_part}" and ``\emph{similar\_to}" are also two kinds of relations of strong and weak hierarchy, but the degree distribution of them is narrow. For relations with strong hierarchy, with the increase of entity degree, the curvature changes with certain regularity. For example, ``\emph{member\_of\_domain\_region}" shows obvious growth and ``\emph{has\_part}" shows stability. But for ``\emph{verb\_group}" and ``\emph{similar\_to}" with weak hierarchy, the curvature changes without regularity. For a complete N-tree, a larger degree corresponds to a larger curvature. In knowledge graph, strong hierarchical relations, such as ``\emph{member\_of\_domain\_region}" and ``\emph{has\_part}", conform to this law to a certain extent. However, this hypothesis is not met for relations with weak hierarchy, because the entities under that relations may present other structures than tree.

\subsubsection{Effect of different curvatures on MRR}
\begin{figure}[htb]
\centering 
\setlength{\belowcaptionskip}{0.1cm}
\setlength{\abovecaptionskip}{0.1cm}
\includegraphics[width=3.5in]{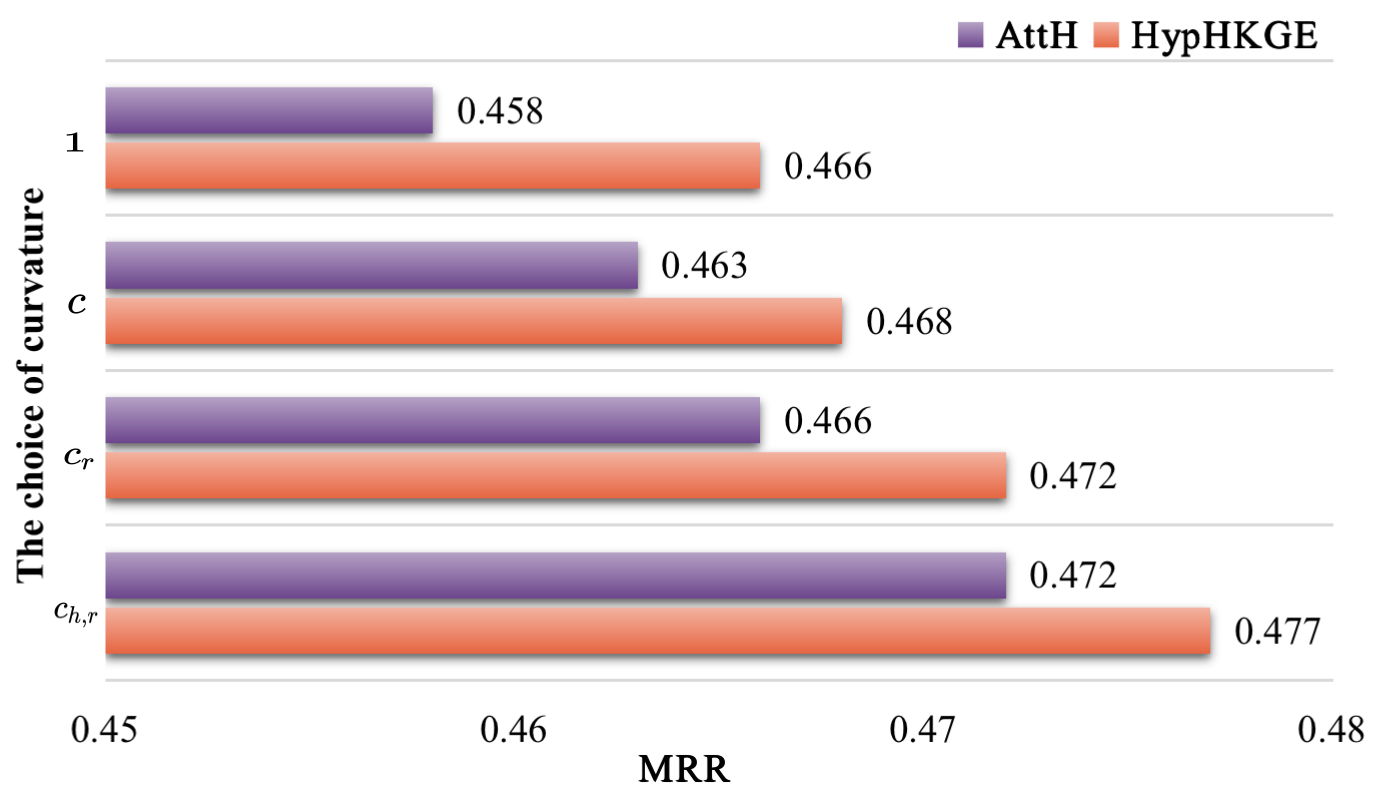} 
\caption{Comparison the MRR of the HypHKGE and the AttH under different curvature choices on WN18RR.} 
\label{fig_5} 	
\end{figure}

The advantage of constructing attention-based curvatures is that we can learn low-dimensional hyperbolic embeddings with high quality. Particularly, compared to AttH that sets specific-relation curvatures (called $c_r$), we model attention-based learnable curvatures (called $c_{h,r}$) that jointly influenced by head entities and relations, thus, various semantic hierarchical structures in KGs are well taken into account. As can be seen from Fig.~\ref{fig_5}, setting the curvature of AttH as $c_{h,r}$, the performance rises by 1.3\% compared to the original, indicating that not only relations but also head entities play a non-negligible role in modeling of curvatures. Also, when setting the same curvature, such as 1, $c$ (all relations share a common curvature) or $c_r$, we find that HypHKGE is also superior to AttH, which fully illustrates the importance of modeling and inferring semantic hierarchies. 

\section{Conclusion}
\label{sec6}
To model semantic hierarchies of KGs in hyperbolic space, we propose the hyperbolic hierarchical knowledge graph embeddings (HypHKGE). The approach constructs attention-based learnable curvatures by considering various hierarchical structures, which play an important role in learning high-quality hyperbolic KGE. Moreover, to infer missing links of KGs, we propose the hyperbolic hierarchical transformations (both inter-level and intra-level transformations) and prove their reasonability. Experiments demonstrate the feasibility and reliability of HypHKGE in capturing the semantic hierarchy in hyperbolic space.

\section{Acknowledgments}
This work is jointly supported by the National Natural Science Foundation of China (No.61876001), and the Natural Science Foundation of Anhui Province (No.1808085MF175). Moreover, we thank the reviewers for their constructive comments and suggestions.

\bibliographystyle{ACM-Reference-Format}
\bibliography{sample-base}

\appendix

\section{Proofs of Results}

\subsection{Hyperbolic Inter-level Transformations}

\label{app1}
$\mathbf{Proposition}$ $\mathbf{1}$: While keeping the direction of the embedding vector constant, the level of $h$ in hyperbolic space becomes $k$ times of the original one, which is the sufficient and necessary condition for the level of $h$ in the tangent space that becomes $k$ times of the original one. That is, it is proved that:
\begin{equation}
	\ \hslash^{c}_{\mathbb{B}}(h_{e})  =k\hslash^{c}_{\mathbb{B}}(h)  \Leftrightarrow \hslash_{\mathbb{E}} ( h_{e})  =k\hslash_{\mathbb{E}}(h)  \\
\end{equation}
\emph{Proof}. First, using Eq.~\ref{eq3}, Eq.~\ref{eq4} and Eq.~\ref{eq4.1}, we know that
\begin{equation}
\label{eqa1}
\begin{aligned}
	\ \hslash^{c}_{\mathbb{B}}(h) &= d_{\mathbb{B}}^{c}(\mathbf{h^H},\mathbf{0}) \\
	 &= \frac{2}{\sqrt{c} } \mathrm{arctanh}( \sqrt{c} \| -\mathbf{h^H}\oplus^{c} \mathbf{0}\|)  \\
	 &=  \frac{2}{\sqrt{c} } \mathrm{arctanh}( \sqrt{c} \| -\mathbf{h^H}\|)
\end{aligned}
\end{equation}
According to Eq.~\ref{eq2}, we also have
\begin{equation}
\label{eqa2}
\mathbf{h^H} = \exp_{\mathbf{0}}^{c}(\mathbf{h^E}) = \frac{\tanh( \sqrt{c} \| \mathbf{h^E}\| )}{\sqrt{c} \| \mathbf{h^E}\| }\mathbf{h^E} 
\end{equation}
Substituting Eq.~\ref{eqa2} into Eq.~\ref{eqa1}, we get a result.
\begin{equation}
\label{eqapp_1}
\begin{aligned}
\ \hslash^{c}_{\mathbb{B}}(h) &= \frac{2}{\sqrt{c} } \mathrm{arctanh}( \sqrt{c} \| -\frac{\tanh( \sqrt{c} \| \mathbf{h^E}\| )}{\sqrt{c} \| \mathbf{h^E}\| }\mathbf{h^E}\|) \\
		 &= 2\|\mathbf{h^E}\|
\end{aligned}
\end{equation}
Similarly, we get:
\begin{equation}
\label{eqapp_2}
	\ \hslash^{c}_{\mathbb{B}}(h_e) = d_{\mathbb{B}}^{c}(\mathbf{h^H},\mathbf{0}) = 2\|\mathbf{h}^{\mathbf E}_e\|
\end{equation}
So, 1) Deriving from the left to the right. Assuming that $\hslash^{c}_{\mathbb{B}}(h_{e})  =k\hslash^{c}_{\mathbb{B}}(h) $, and according to Eq.~\ref{eqapp_1} and Eq.~\ref{eqapp_2}, we get:
\begin{equation}
	\|\mathbf{h}^{\mathbf E}_e\|=k\|\mathbf{h^E}\|
\end{equation}
Also since:
\begin{equation}
\label{eqapp_3}
	\ \|\mathbf{h}^{\mathbf E}_e\| = d_{\mathbb{E}}(\mathbf{h}^{\mathbf E}_e,\mathbf{0}) = \hslash_{\mathbb{E}} ( h_{e})
\end{equation}
\begin{equation}
\label{eqapp_4}
	\ \|\mathbf{h^E}\| = d_{\mathbb{E}}(\mathbf{h^E},\mathbf{0}) = \hslash_{\mathbb{E}} (h)
\end{equation}
Finally, we get:
\begin{equation}
	\ \hslash_{\mathbb{E}} ( h_{e})  =k\hslash_{\mathbb{E}}(h)  
\end{equation}
On the contrary, 2) Deriving from the right to the left. Assuming that $\hslash_{\mathbb{E}}(h_{e})  =k\hslash_{\mathbb{E}}(h) $, and according to Eq.~\ref{eqapp_3} and Eq.~\ref{eqapp_4}, we get:
\begin{equation}
	\|\mathbf{h}^{\mathbf E}_e\|=k\|\mathbf{h^E}\|
\end{equation}
Finally, according to Eq.~\ref{eqapp_1} and Eq.~\ref{eqapp_2}, we get:
\begin{equation}
	\ \hslash^{c}_{\mathbb{B}}(h_{e})  =k\hslash^{c}_{\mathbb{B}}(h)
\end{equation}

\subsection{Hyperbolic Intra-level Transformations}

\label{app2}
$\mathbf{Proposition}$ $\mathbf{2}$: For entity $h$, given the hyperbolic embeddings $\mathbf{h^H}$ and the tangent embeddings $\mathbf{h^E}$, using the intra-level transformations factor $\Theta_r$ in hyperbolic space have the same effect as using in tangent space. That is, given $\mathbf{h^H}\in \mathbb{B}^{d, c}$ and $ \mathbf{h^E} = \log_{\mathbf{0}}^{c}(\mathbf{h^H})$, we conclude that:
\begin{equation}
\label{eqapp2_1}
	\ \exp_{\mathbf{0}}^{c}(\Theta_r\mathbf{h^E}) =\Theta_r\mathbf{h^H}
\end{equation}
\emph{Proof}. First, the left side of the equation:
\begin{equation}
	\ \exp_{\mathbf{0}}^{c}(\Theta_r\mathbf{h^E}) = \frac{\tanh( \sqrt{c} \| \Theta_r\mathbf{h^E}\| )}{\sqrt{c} \| \Theta_r\mathbf{h^E}\| }\Theta_r\mathbf{h^E}
\end{equation}
since the rotation does not change the L2 norm, i.e., $\| \Theta_r\mathbf{h^E}\|=\| \mathbf{h^E}\|$, we get:
\begin{equation}
	\ \exp_{\mathbf{0}}^{c}(\Theta_r\mathbf{h^E}) = \frac{\tanh( \sqrt{c} \| \mathbf{h^E}\| )}{\sqrt{c} \| \mathbf{h^E}\| }\Theta_r\mathbf{h^E}
\end{equation}
where $\frac{\tanh( \sqrt{c} \| \Theta_r\mathbf{h^E}\| )}{\sqrt{c} \| \Theta_r\mathbf{h^E}\| }$ is a scalar. Next, the right side of Eq.~\ref{eqapp2_1}:
\begin{equation}
\begin{aligned}
	\ \Theta_r\mathbf{h^H} &= \Theta_r\exp_{\mathbf{0}}^{c}(\mathbf{h^E}) \\
	&= \frac{\tanh( \sqrt{c} \| \mathbf{h^E}\| )}{\sqrt{c} \| \mathbf{h^E}\| }\Theta_r\mathbf{h^E}
\end{aligned}
\end{equation}
Therefore, the left side of Eq.~\ref{eqapp2_1} is equal to the right.

\end{document}